\documentclass[runningheads]{llncs}
\usepackage{graphicx}
\usepackage{comment}
\usepackage{amsmath,amssymb}
\usepackage{color}
\usepackage{graphicx}
\usepackage{subfigure}
\usepackage{booktabs}
\usepackage{float}

\newcommand{\specialcell}[2][c]{%
  \begin{tabular}[#1]{@{}c@{}}#2\end{tabular}}
\newcommand{\etal}{\textit{et al.}}
\usepackage{algorithm,algpseudocode}
\usepackage{float}
\usepackage{amssymb}
\usepackage{colorspace}
\definespotcolor{mygreen}{PANTONE 7716 C}{.83, 0, .40, .11}

\begin{document}

\pagestyle{headings}
\mainmatter

\title{Learning Surrogates via Deep Embedding}

\titlerunning{Learning Surrogates via Deep Embedding}

\author{Yash Patel, Tom{\'a}{\v{s}} Hoda{\v{n}}, Ji{\v{r}}{\'i} Matas}

\authorrunning{Y. Patel~\etal}
\institute{Visual Recognition Group, Czech Technical University in Prague \\
\email{\{patelyas,hodanto2,matas\}@fel.cvut.cz}
}

\maketitle

\begin{abstract}

This paper proposes a technique for training a neural network by minimizing a surrogate loss that approximates the target evaluation metric, which may be non-differentiable. The surrogate is learned via a deep embedding where the Euclidean distance between the prediction and the ground truth corresponds to the value of the evaluation metric. The effectiveness of the proposed technique is demonstrated in a post-tuning setup, where a trained model is tuned using the learned surrogate. Without a significant computational overhead and any bells and whistles, improvements are demonstrated on challenging and practical tasks of scene-text recognition and detection. In the recognition task, the model is tuned using a surrogate approximating the edit distance metric and achieves up to $39\%$ relative improvement in the total edit distance. In the detection task, the surrogate approximates the intersection over union metric for rotated bounding boxes and yields up to $4.25\%$ relative improvement in the $F_{1}$ score.

\end{abstract}

\section{Introduction}
\label{sec:introduction}

Supervised learning of a neural network involves minimizing a differentiable loss function on annotated data. The differentiable nature of the loss function and the network architecture allows the model weights to be updated via backpropagation \cite{rumelhart1986learning}. The performance on a wide range of computer vision tasks have significantly improved thanks to the progress in deep neural network architectures \cite{krizhevsky2012imagenet,he2016deep,simonyan2014very} and the introduction of large scale supervised datasets~\cite{deng2009imagenet,lin2014microsoft}. As designing architectures often demands detailed domain expertise and creating new datasets is expensive, there has been a substantial effort in automating the process of designing better task-specific architectures \cite{elsken2018neural,ryoo2019assemblenet,zoph2016neural} and employing self-supervised methods of learning to reduce the dependence on human-annotated data \cite{gidaris2018unsupervised,caron2018deep,gomez2017self}. However, little attention has been paid to automate the process of designing the loss functions. 

For many practical problems in computer vision, models are trained with simple proxy losses, which may not align with the evaluation metric. The evaluation metric may not always be differentiable, prohibiting its use as a loss function. An example of a non-differentiable metric is the visible surface discrepancy (VSD)~\cite{hodan2018bop} used to evaluate 6D object pose estimation methods. Another example is the edit distance (ED) defined by counting unit operations (addition, deletion, and substitution) necessary to transform one text string into another and is a common choice for evaluating scene text recognition methods \cite{karatzas2015icdar,karatzas2013icdar,nayef2019icdar2019}. Since ED is non-differentiable, the methods use either CTC \cite{graves2006connectionist} or per-character cross-entropy~\cite{baek2019wrong} as the proxy loss. Yet another popular non-differentiable metric is the intersection over union (IoU) used to compare the predicted and the ground truth bounding boxes when evaluating object detection methods. Although these methods typically resort to using proxy losses such as {\em smooth-L$_{1}$} \cite{ren2015faster} or {\em L$_{2}$} \cite{redmon2016you}, Rezatofighi~\etal~\cite{rezatofighi2019generalized} demonstrate that there is no strong correlation between {\em L$_{n}$} objectives and IoU. Further, Yu~\etal~\cite{yu2016unitbox} show that IoU accounts for a bounding box as a whole whereas regressing using an {\em L$_{n}$} proxy loss treats each point independently.

\begin{figure}[t!]
    \centering
    \includegraphics[width=0.9\textwidth, trim={2.7cm 1.7cm 1cm 0cm},clip]{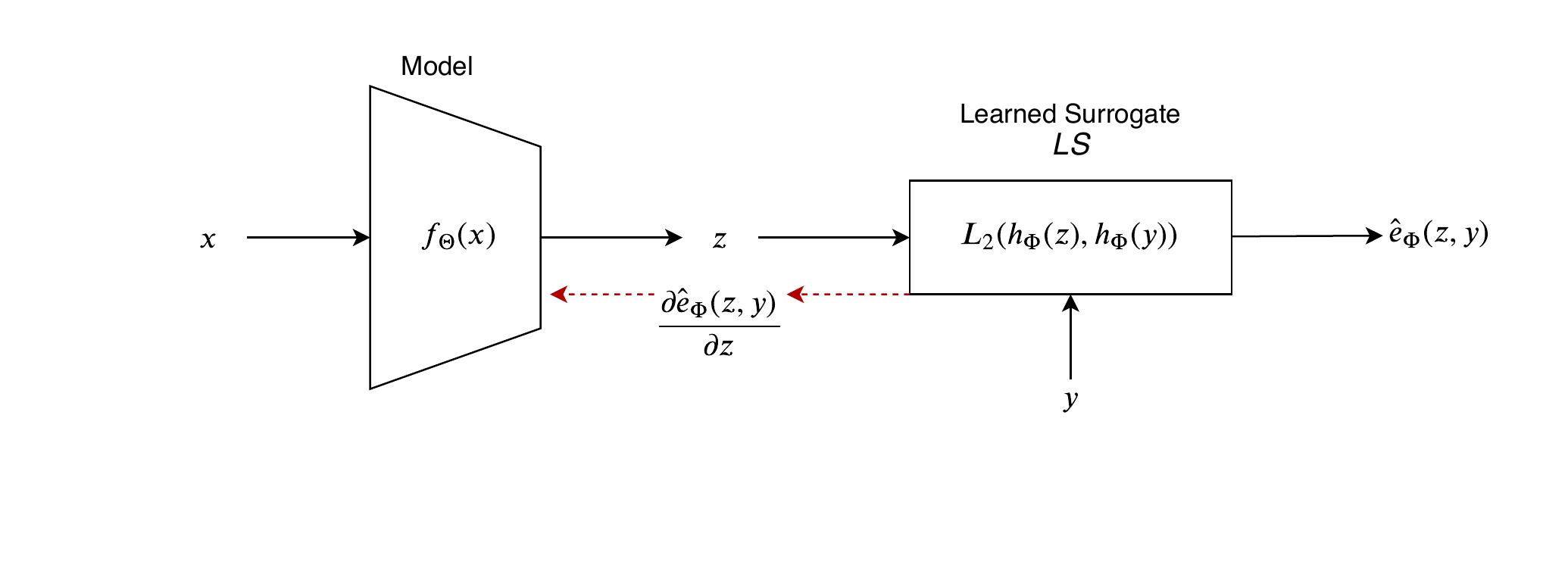}
    \caption{For the input $x$ with the corresponding ground-truth $y$, the model being trained outputs $z=f_{\Theta}(x)$. The learned surrogate provides a differentiable approximation of the evaluation metric: $\hat{e}_{\Phi}(z,y)=L_{2}(h_{\Phi}(z), h_{\Phi}(y))$,
    where $h_{\Phi}$ is a learned deep embedding model, and $h_{\Phi}(z)$ and $h_{\Phi}(y)$ are embedding representations for the prediction and the ground truth, respectively. Model $f_{\Theta}(x)$ for the target task (\emph{e.g.} scene text recognition or detection) is trained with the gradients from the surrogate: $\frac{\partial(\hat{e}_{\Phi}(z,y))}{\partial z}$.}
    \label{fig:overall_lsl}
\end{figure}

For popular metrics such as IoU, hand-crafted differentiable approximations have been designed \cite{yu2016unitbox,rezatofighi2019generalized}. However, hand-crafting a surrogate is not scalable as it requires domain expertise and may involve task-specific assumptions and simplifications. The IoU-loss introduced in \cite{yu2016unitbox,rezatofighi2019generalized} allows for optimization on the evaluation metric directly but makes a strong assumption about the bounding boxes to be axis-aligned. In numerous practical applications such as aerial image object detection~\cite{xia2018dota}, scene text detection~\cite{karatzas2015icdar} and visual object tracking~\cite{kristan2019seventh}, the bounding boxes may be rotated and the methods for such tasks revert to using simple but non-optimal proxy loss functions such as {\em smooth-L$_{1}$}~\cite{ma2018arbitrary,buvsta2018e2e,azimi2018towards}.

To address the aforementioned issues, this paper proposes to learn a differentiable surrogate that approximates the evaluation metric and use the learned surrogate to optimize the model for the target task. The metric is approximated via a deep embedding, where the Euclidean distance between the prediction and the ground truth corresponds to the value of the metric. The mapping to the embedding space is realized by a neural network, which is learned using only the value of the metric. Gradients of this value with respect to the inputs are not required for learning the surrogate. In fact, the gradients may not even exist, as is the case of the edit distance metric. Throughout this paper, we refer to the proposed method for training with learned surrogates as ``LS". Figure \ref{fig:overall_lsl} provides an overview of the proposed method.

In this paper, the focus on a post-tuning setup, where a model that has converged on a proxy loss is tuned with LS. We consider two different optimization tasks: post-tuning with a learned surrogate for the edit distance (LS-ED) and the IoU of rotated bounding boxes (LS-IoU). To the best of our knowledge, we are the first to optimize directly on these evaluation metrics.

The rest of the paper is structured as follows. Related work is reviewed in Section~\ref{sec:related_work}, the technique for learning the surrogate and training with it is presented in Section~\ref{sec:lsl}, experiments are shown in Section~\ref{sec:experiments} and the paper is concluded in Section~\ref{sec:conclusion}.

\section{Related Work}
\label{sec:related_work}

Training machine learning models by directly minimizing the evaluation metric has been shown effective on various tasks. For example, the state-of-the-art learned image compression \cite{lee2018context,balle2018variational} and super-resolution \cite{ledig2017photo,dong2015image} methods directly optimize the perceptual similarity metrics such as MS-SSIM \cite{wang2003multiscale} and the peak signal-to-noise ratio (PSNR). Certain compression methods optimize on an approximate of human perceptual similarity, which is learned in a supervised manner using annotated data \cite{patel2019deep,patel2020hierarchical}. Image classification methods \cite{krizhevsky2012imagenet,he2016deep,simonyan2014very} are typically trained with the cross-entropy loss that has been shown to align well with the misclassification rate, \emph{i.e.} the evaluation metric, under the assumption of large scale and clean data~\cite{berrada2018smooth,lapin2016loss}.

When designing evaluation metrics for practical computer vision tasks, the primary goal is to fulfil the requirements of potential applications and not to ensure the metrics being amenable to an optimization approach. As a consequence, many evaluation metrics are non-differentiable and cannot be directly minimized by the currently popular gradient-descent optimization approaches. For example, the visible surface discrepancy~\cite{hodan2018bop}, which is used to evaluate 6D object pose estimation methods, was designed to be invariant under pose ambiguity. This is achieved by calculating the error only over the visible part of the object surface, which requires a visibility test that makes the metric non-differentiable. Another example is the edit distance metric~\cite{karatzas2015icdar,DBLP:conf/icdar/GomezSGNVMBK17}, which is used to evaluate scene text recognition methods and is calculated via dynamic programming, which makes it infeasible to obtain the gradients.

There have been efforts towards approximating non-differentiable operations in a differentiable manner to enable end-to-end training. Kato~\etal~\cite{kato2018neural} proposed a neural network to approximate rasterization, allowing for a direct optimization on IoU for 3D reconstruction. Agustsson~\etal~\cite{agustsson2017soft} proposed a soft-to-hard vector quantization mechanism. It is based on soft cluster assignments during backpropagation, which allows neural networks to learn tasks involving quantization, \emph{e.g.} the image compression. Our work differs as we propose a general approach to approximate the evaluation metric, instead of approximating task-specific building blocks of neural networks.

Another line of research has focused on hand-crafting differentiable approximates of the evaluation metrics, which either align better with the metrics or enable training on them directly. Prabhavalkar~\etal~\cite{prabhavalkar2018minimum} proposed a way of optimizing attention based speech recognition models directly on word error rate. As mentioned earlier, \cite{yu2016unitbox,rezatofighi2019generalized} proposed ways for directly optimizing on intersection-over-union (IoU) as the loss for the case of axis-aligned bounding boxes.  Rahman~\etal~\cite{rahman2016optimizing} proposed a hand-crafted approximation of IoU for semantic segmentation.

Learning task-specific surrogates has been attempted. Nagendra~\etal~\cite{nagendar2018neuro} demonstrated that learning the approximate of IoU leads to better performance in the case of semantic segmentation. However, the method requires custom operations to estimate true and false positives, and false negatives, which makes the learning approach task-specific. Engilberge~\etal~\cite{DBLP:conf/cvpr/EngilbergeCPC19} proposed a learned surrogate for sorting-based tasks such as cross-modal retrieval, multi-label image classification and visual memorability ranking. Their results on sorting-based tasks suggest that learning the loss function could outperform hand-crafted losses.

More closely related to our work is the direct loss method by Hazan~\etal~\cite{hazan2010direct} where a surrogate loss is minimized by embedding the true loss as a correction term. Song~\etal~\cite{song2016training} extended this approach to the training of neural networks. However, it assumes that the loss can be disentangled into per-instance sub-losses, which is not always feasible, \emph{e.g.} the $F_{1}$ score \cite{grabocka2019learning} involves two non-decomposable functions (recall and precision). An alternative is to directly learn the amount of update values that are applied to the parameters of the prediction model. The framework proposed in \cite{li2017learning} includes a controller that uses per-parameter learning curves comprised of the loss values and derivatives of the loss with respect to each parameter. The method suffers from two drawbacks that prohibit its direct application to training on evaluation metrics: a) for large networks, it is computationally infeasible to store the learning curve of every parameter, and b) no gradient information is available for non-differentiable losses.

Our work is similar to the approach by Grabocka~\etal~\cite{grabocka2019learning}, where the evaluation metric is approximated by a neural network. Their approach differs as the network learning the surrogate takes both the prediction and the ground truth as the input and directly regresses the value of the metric. Since we formulate the task as embedding learning and train the surrogate such that the $L_{2}$ in the embedded space corresponds to the metric, our method ensures that the gradients are smaller when the prediction is closer to the ground truth. Furthermore, as illustrated in Section \ref{sec:lsl}, we learn the surrogate with an additional gradient penalty term to ensure that the gradients obtained from our learned surrogate are bounded for stable training.

\section{Learning Surrogates via Deep Embedding}
\label{sec:lsl}

Say that the supervised task is being learned from samples drawn uniformly from a distribution $(x,y)\sim P_{D}$. For a given input $x$ and an expected output $y$, a neural network model outputs $z=f_{\Theta}(x)$ where $\Theta$ are the model parameters learned via backpropagation as:

\begin{equation}
    \Theta_{t+1} \leftarrow \Theta_{t} - \eta \frac{\partial l(z,y)}{\partial \Theta_{t}}
\end{equation}
where $l(z,y)$ is a differentiable loss function, $t$ is the training iteration, and $\eta$ is the learning rate.

The model trained with loss $l(z,y)$ is evaluated using metric $e(z,y)$. When metric $e(z,y)$ is differentiable, it can be directly used as the loss. The technique proposed in this paper addresses the cases when metric $e(z,y)$ is non-differentiable by learning a differentiable surrogate loss denoted as $\hat{e}_{\Phi}(z,y)$. The learned surrogate is realized by a neural network, which is differentiable and is used to optimize the model. The weight updates are:

\begin{equation}
\Theta_{t+1} \leftarrow \Theta_{t} - \eta \frac{\partial \hat{e}_{\Phi}(z,y)}{\partial \Theta_{t}}
\end{equation}

\subsection{Definition of the Surrogate}

The surrogate is defined via a learned deep embedding $h_{\Phi}$ where the Euclidean distance between the prediction $z$ and the ground truth $y$ corresponds to the value of the evaluation metric:

\begin{equation}
    \hat{e}_{\Phi}(z,y) = \left\Vert h_{\Phi}(z) - h_{\Phi}(y)\right\Vert_{2}
    \label{eq:defination_ehat}
\end{equation}

\subsection{Learning the Surrogate}
\label{sec:learning_the_surrogate}

Learning the surrogate, \emph{i.e.} approximating the evaluation metric, with a deep neural network is formulated as a supervised learning task requiring three major components: a model architecture, a loss function, and a source of training data.

\subsubsection{Architecture.}
In this paper, the architecture is designed manually, such that it is suitable for the nature of the inputs $z$ and $y$ (details are in Section~\ref{sec:experiments}). Modern approaches for architecture search, \emph{e.g.}~\cite{elsken2018neural,ryoo2019assemblenet,zoph2016neural}, could yield better results but are computationally expensive.

\subsubsection{Training Loss.}

The surrogate is learned with the following objectives:
\begin{enumerate}
    \item The learned surrogate corresponds to the value of the evaluation metric: \begin{equation}
        \hat{e}_{\Phi}(z,y) \approx e(z,y)
    \end{equation}
    \item The first order derivative of the learned surrogate with respect to the prediction $z$ is close to $1$: \begin{equation}
       \left\Vert \frac{\partial \hat{e}_{\Phi}(z,y)}{\partial z} \right\Vert_{2} \approx 1
       \label{eq:gp_term}
    \end{equation}
\end{enumerate}

Both objectives are realized and linearly combined in the training loss:
\begin{equation}
    \text{loss}(z,y) = \left\Vert \big(\hat{e}_{\Phi}(z,y) - e(z,y)\right\Vert_2^2 + \lambda \left(\left\Vert \frac{\partial \hat{e}_{\Phi}(z,y)}{\partial z} \right\Vert_{2} - 1\right)^{2}
    \label{eq:lsl_loss}
\end{equation}

Bounding the gradients (Equation~\ref{eq:gp_term}) has shown to enhance the training stability for Generative Adversarial Networks~\cite{gulrajani2017improved} and has shown to be useful for learning the surrogate. Parameters $\Phi$ of the embedding model $h_\Phi$ are learned by minimizing the loss (Equation~\ref{eq:lsl_loss}).

\subsubsection{Source of Training Data.}
Source of the training data for learning the surrogate determines the quality of the approximation over the domain. The model $f_{\Theta}(x) = z$ for the supervised task is trained on samples obtained from a dataset $D$. Let us assume that $R$ is a random data generator providing examples for the learning of the surrogate, sampled uniformly in the range of the evaluation metric (see Section~\ref{sec:experiments} for details). Note that $R$ is independent of $f_{\Theta}(x)$.

Three possibilities for the data source are considered:
\begin{enumerate}
    \item \textit{Global approximation}: $(z,y) \sim P_{R}$.
    \item \textit{Local approximation}: $(z,y) \sim P_{f_{\Theta}(x)}$, where $(x,y) \sim P_{D}$.
    \item \textit{Local-global approximation}: $(z,y) \sim P_{f_{\Theta}(x) \cup R}$.
\end{enumerate}

The local-global approximation yields a high quality of both the approximation and gradients (Section \ref{sec:exp_analysis}) and is therefore used in the main experiments.

\subsection{Training with the Learned Surrogate}
\label{sec:training_with_the_learned_surrogate}

The learned surrogate is used in a post-tuning setup, where model $f_{\Theta}(x)$ has been pre-trained using a proxy loss. This setup ensures that $f_{\Theta}(x)$ is not generating random outputs and thus simplifies post-tuning with the surrogate. The parameters of the surrogate $\Phi$ are initialized randomly.

Learning of the surrogate $\hat{e}_{\Phi}$ and post-tuning of the model $f_{\Theta}(x)$ are conducted alternatively. The surrogate parameters $\Phi$ are updated first while the model parameters $\Theta$ are fixed. The surrogate is learned by sampling $(z,y)$ jointly from the model and the random generator. Subsequently, the model parameters are trained while the surrogate parameters are fixed. Algorithm \ref{alg:train_with_lsl} demonstrates the overall training procedure.

\begin{algorithm}
\caption{Training with LS \textit{(local-global approximation)}}
\label{alg:train_with_lsl}
\textbf{Inputs}: Supervised data $D$, random data generator $R$, evaluation metric $e$.\\
\textbf{Hyper-parameters}: Number of update steps $I_{a}$ and $I_{b}$, learning rates $\eta_{a}$ and $\eta_{b}$, number of epochs $E$.\\
\textbf{Objective}: Train the model for a given task that is $f_{\Theta}(x)$ and the surrogate ,\emph{i.e.}, $e_{\Phi}$.
\begin{algorithmic}[1]
\State \textit{Initialize} $\Theta \leftarrow$ pre-trained weights, $\Phi \leftarrow$ random weights.
\For{epoch = 1,...,E}
    \For{i = 1,...,$I_{a}$}
        \State sample $(x,y) \sim P_{D}$, sample $(z_{r},y_{r}) \sim P_{R}$
        \State inference $z=f_{\Theta^{epoch-1}}(x)$
        \State compute loss $l_{\hat{e}} =$ {\em loss}$(z,y) + ${\em loss}$(z_{r},y_{r})$ (Equation \ref{eq:lsl_loss})
        \State $\Phi^{i} \leftarrow \Phi^{i-1} - \eta_{a} \frac{\partial l_{\hat{e}}}{\partial \Phi^{i-1}}$
    \EndFor \\
   \quad \quad $\Phi \leftarrow \Phi^{I_{a}}$
    \For{i = 1,...,$I_{b}$}
        \State sample $(x,y) \sim P_{D}$
        \State inference $z = f_{\Theta^{i-1}}(x)$
        \State compute loss $l_{f} = \hat{e}_{\Phi^{epoch}}(z,y)$ (Equation \ref{eq:defination_ehat})
        \State $\Theta^{i} \leftarrow \Theta^{i-1} - \eta_{b} \frac{\partial(l_{f})}{\partial \Theta^{i-1}}$
    \EndFor \\
    \quad \quad $\Theta \leftarrow \Theta^{I_{b}}$
\EndFor
\end{algorithmic}
\end{algorithm}

\section{Experiments}
\label{sec:experiments}

The efficacy of LS is demonstrated on two different tasks: post-tuning with a learned surrogate for the edit distance (Section \ref{sec:exp_ed}) and for the IoU of rotated bounding boxes (Section \ref{sec:exp_iou}). This section provides details of the models for these tasks, design choices for learning the surrogates and empirical evidence showing the efficacy of LS. Unless stated otherwise, the results were obtained using the local-global approximation setup as elaborated in Algorithm \ref{alg:train_with_lsl}.

\subsection{Analysing the Learned Surrogates} \label{sec:exp_analysis}

The aspects considered for evaluating the surrogates are:
\begin{enumerate}
    \item The quality of approximation $\hat{e}_{\Phi}(z,y)$.
    \item The quality of gradients $\frac{\partial(\hat{e}_{\Phi}(z,y))}{\partial z}$.
\end{enumerate}

Both the quality of the approximation and the gradients depend on three components: an architecture, a loss function, and a source of training data (Section \ref{sec:learning_the_surrogate}). Given an architecture, the choices for the loss function to learn the surrogate and the training data are justified subsequently.

\subsubsection{Quality of approximation.} 
The quality of the approximation is judged by comparing the value of the surrogate with the value of the evaluation metric, calculated on samples obtained from model $f_{\Theta}(x)$. When learning the surrogate, higher quality of approximation is enforced by the mean squared loss between $e(z,y)$ and $\hat{e}_{\Phi}(z,y)$ (the first term on the right-hand side of Equation \ref{eq:lsl_loss}). Figure \ref{fig:ls_ed_plots} (left) shows the quality of the approximation measured by the $L_{1}$ distance between the learned surrogate and the edit distance. It can be seen that the surrogate approximates the edit distance accurately (the $L_{1}$ distance drops swiftly below $0.2$, which is negligible for the edit distance).

\subsubsection{Quality of gradients.} 
Judging the quality of gradients is more complicated. When learning the surrogate, the gradient-penalty term attempts to make the gradients bounded, \emph{i.e.} to make the training stable (second term on the right-hand side of the equation \ref{eq:lsl_loss}). However, this is not sufficient if the gradients do not optimize $f_{\Theta}(x)$ on the evaluation metric. We rely on the improvement or the decline in the performance of the model $f_{\Theta}(x)$ to judge the quality of the gradients. Table \ref{table:detection_results} shows that the local-global approximation leads to the largest improvements when optimizing on IoU for rotated bounding boxes.

\subsubsection{Choice of training data.}
\begin{figure}[t!]
    \centering
    \includegraphics[width=\textwidth,trim={0cm 0.5cm 0cm 0cm},clip]{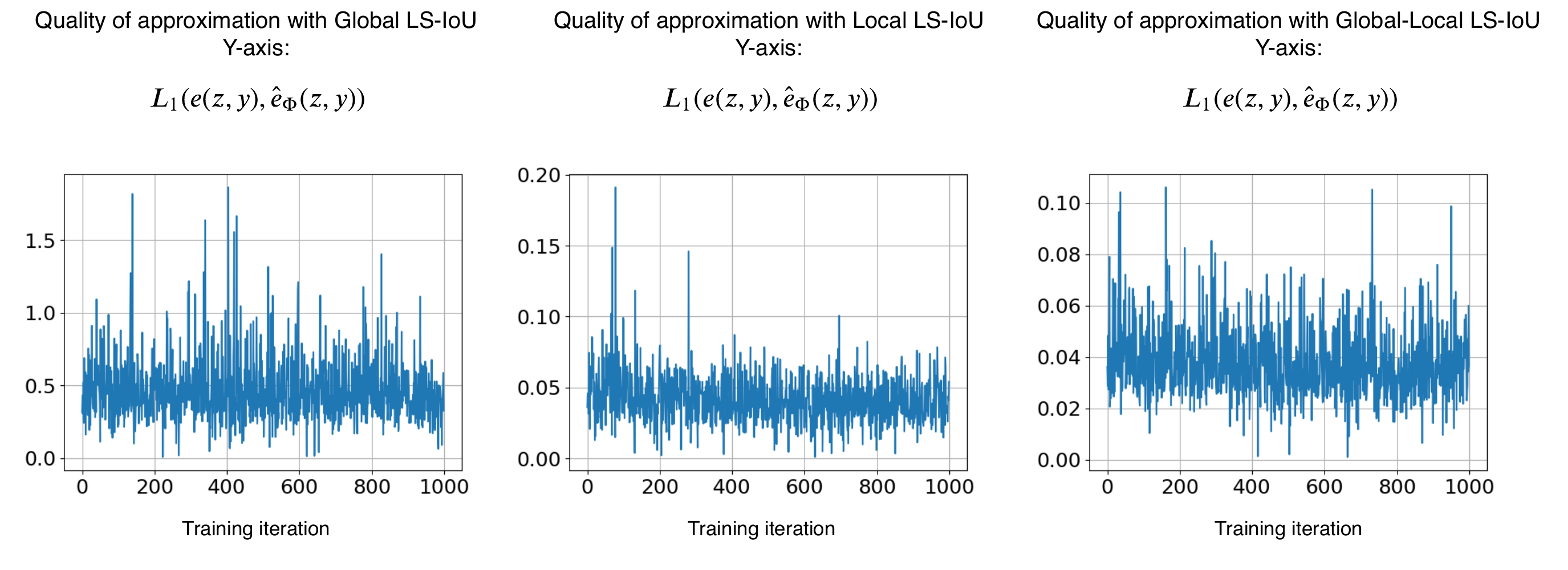}
    \caption{The error in the approximation of the IoU for rotated bounding boxes is shown for the first $1K$ iterations of the training with LS-IoU. Error is measured by the $L_{1}$ distance between IoU and the surrogate. It can be seen that the error is high for the global and low for the local and global-local approximation variants.}
    \label{fig:ls_iou_plots}
\end{figure}

Figure \ref{fig:ls_iou_plots} shows the quality of approximation with different choices of training data for learning the surrogate. These empirical observations suggest that using global approximation leads to a low quality of the approximation. This can be accounted to the domain gap between the data obtained from the random generator and the model. Using the local approximation leads to a higher quality of the approximation, however, the gradients obtained from the surrogate are not useful to train $f_{\Theta}(x)$ (Table. \ref{table:detection_results}), \emph{i.e.} although the quality of the approximation is high, the quality of gradients is not. This can be attributed to surrogate over-fitting on samples obtained from the model and losing generalization capability on samples outside this distribution. Finally, it was observed that using the local-global approximation leads to both properties -- high quality of approximation and high quality of gradients.

\begin{figure}[t!]
    \centering
    \includegraphics[width=0.8\textwidth,trim={1cm 0cm 1cm 0cm},clip]{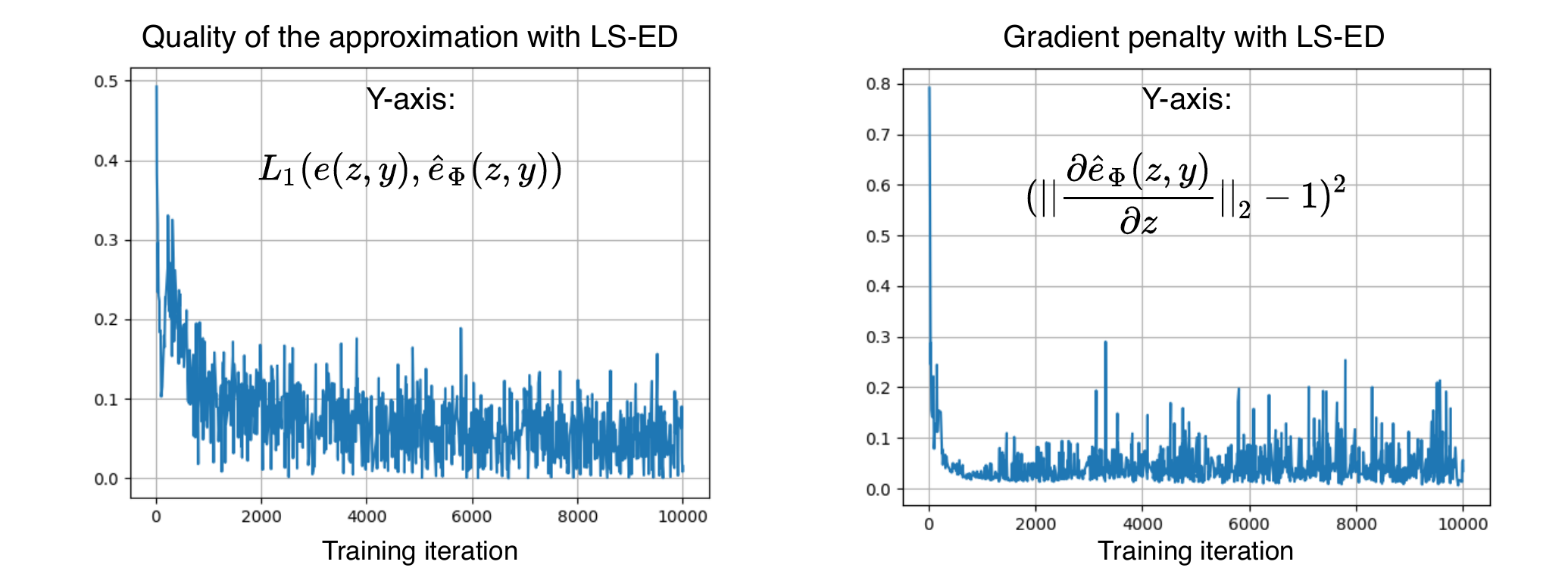}
    \caption{\textbf{Left}: The error in approximation for the first $10K$ training iterations. The error is obtained by computing the $L_{1}$ distance between the true edit distance values and the LS-ED predictions and dividing by the batch size. Note that the edit distance can only take non-negative integer values, thus the error in the range of $0-0.2$ is fairly low. \textbf{Right}: The gradient penalty term from the optimization of the LS-ED model (Equation \ref{eq:lsl_loss}).}
    \label{fig:ls_ed_plots}
\end{figure}

\subsection{Post-Tuning with a Learned Surrogate for ED (LS-ED)}
\label{sec:exp_ed}

It is experimentally shown that LS can improve scene text recognition models (STR) on edit distance (ED), which is a popularly used metric to evaluate STR methods~\cite{karatzas2015icdar,karatzas2013icdar,nayef2019icdar2019}. The empirical evidence shows that post-tuning STR models with LS-ED lead to improved performance on various metrics such as accuracy, normalized edit distance, and total edit distance~\cite{DBLP:conf/icdar/GomezSGNVMBK17}.

\subsubsection{Scene Text Recognition (STR).}
Given an input image of a cropped word, the task of STR is to generate the transcription of the word. The state-of-the-art architectures for scene text recognition can be factorized into four modules \cite{baek2019wrong} (in this order): (a) transformation, (b) feature extraction, (c) sequence modelling, and (d) prediction. The feature extraction and prediction are the core modules of any STR model and are always employed. On the other hand, transformation and sequence modelling are not essential but have shown to improve the performance on benchmark datasets. Post-tuning with LS-ED is investigated for two different configurations of STR models.

The transformation module attempts to rectify the curved or tilted text, making the task easier for the subsequent modules of the model. It is learned jointly with the rest of the modules, and a popular choice is thin-plate spline (TPS) \cite{DBLP:conf/cvpr/ShiWLYB16,DBLP:conf/nips/JaderbergSZK15,DBLP:conf/bmvc/LiuCWSH16}. TPS can be either present or absent in the overall STR model. 

The feature extraction module maps the image or its transformed version to a representation that focuses on the attributes relevant for character recognition, while the irrelevant features are suppressed. Popular choices include VGG-16 \cite{simonyan2014very} and ResNet \cite{he2016deep}. It is a core module of the STR model and is always present. 

The features are the input of the sequence modelling module, which captures the contextual information within a sequence of characters for the next module to predict each character more robustly. BiLSTM \cite{hochreiter1997long} is a popular choice.

The output character sequence is predicted from the identified features of the image. The choice of the prediction module depends on the loss function used for training the STR model. Two popular choices of loss functions are CTC \cite{graves2006connectionist} (sigmoid output) or attention \cite{DBLP:conf/cvpr/ShiWLYB16} (per-character softmax output). 

Baek~\etal~\cite{baek2019wrong} provides a detailed analysis of STR models and the impact of different modules on the performance. Following \cite{baek2019wrong}, LS-ED is investigated with the state-of-the-art performing configuration, which is \textit{TPS-ResNet-BiLSTM-Attn}. To demonstrate the efficacy of LS-ED, results are also shown with \textit{ResNet-BiLSTM-Attn}, \emph{i.e.}, the transformation module is removed. Note that the CTC based prediction has been shown to consistently perform worse compared to the attention counter-part \cite{baek2019wrong}, and thus the analysis in this paper has been narrowed down to only the attention-based prediction.

Similar to \cite{baek2019wrong}, the STR models are trained on the union of the synthetic data obtained from MJSynth \cite{DBLP:journals/corr/JaderbergSVZ14} and SynthText \cite{DBLP:conf/cvpr/GuptaVZ16} resulting in a total of $14.4$ million training examples. Furthermore, following the standard setup of \cite{baek2019wrong}, there is no fine-tuning performed in a dataset-specific manner before the final testing. Let us say that the STR model is $f_{\Theta}(x)$, such that $f_{\Theta}:\mathbb{R}^{100\times 32 \times 1}\xrightarrow{}\mathbb{R}^{|A|\times L}$. The dimensions of the input cropped word image $x$ is fixed to $100\times 32 \times 1$ (gray-scale). The output for attention based prediction module is a per-character softmax over the set of characters. Here $L$ is the maximum length of characters in the word and $|A|$ is the number of characters. During inference, argmax is performed at each character location to output the predicted text string. The ground truth $y$ is represented as a per-character one-hot vector.

The STR models are first trained with the proxy loss, \emph{i.e.}, cross-entropy for $300K$ iterations with a mini-batch size of $192$. The models are optimized using ADADELTA \cite{DBLP:journals/corr/abs-1212-5701} (same setup as \cite{baek2019wrong}). Once the training is completed these models are tuned with LS-ED on the same set of $14.4$ million training examples for another $20K$ iterations. The models trained purely on the synthetic datasets are tested on a collection of real datasets - IIIT-5K \cite{mishra2012scene}, SVT \cite{wang2011end}, ICDAR'03 \cite{lucas2003icdar}, ICDAR'13 \cite{karatzas2013icdar}, ICDAR'15 \cite{karatzas2015icdar}, SVTP \cite{quy2013recognizing}  and CUTE \cite{risnumawan2014robust} datasets. 

\begin{figure}[t!]
    \centering
    \includegraphics[width=\textwidth]{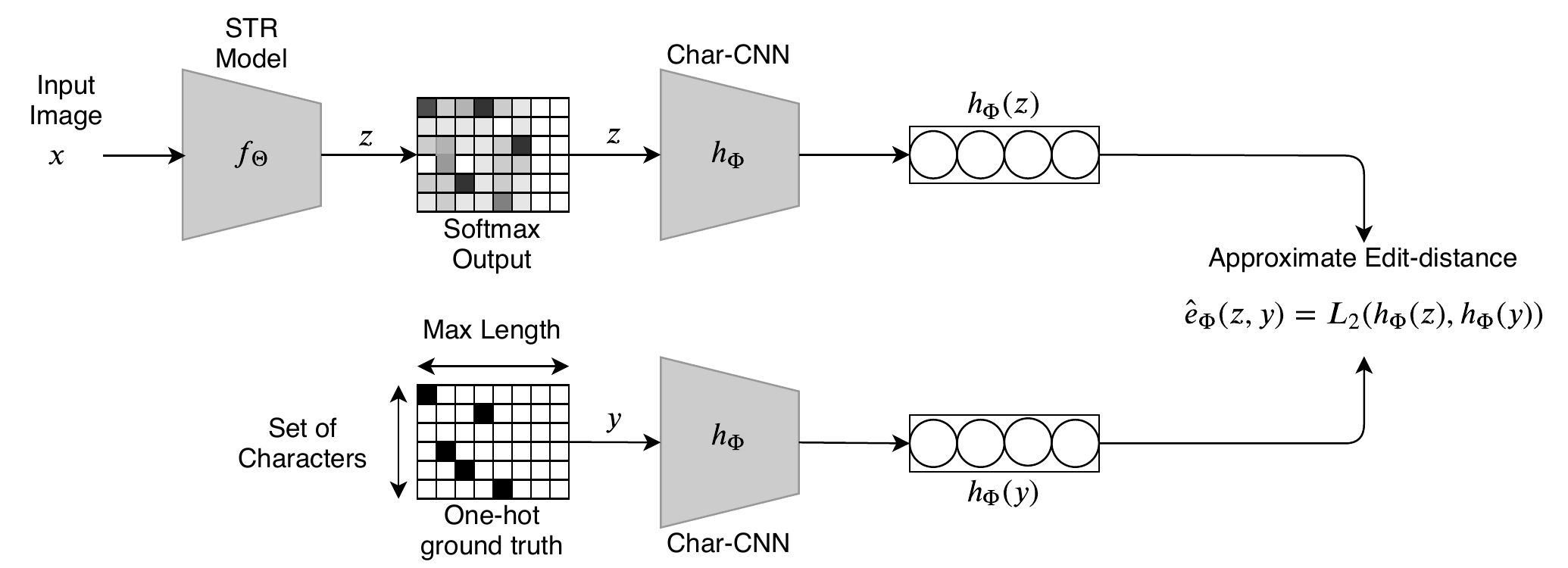}
    \caption{Training scene text recognition (STR) models with LS-ED. The output of the STR model $z_{|A|\times L}$ and the ground-truth $y_{|A|\times L}$ ($L$ is the maximum length of the word and $A$ is the set of characters) are fed to the Char-CNN embedding model to obtain embedding vectors, $h_{\Phi}(z)$ and $h_{\Phi}(y)$ respectively. The approximate edit distance value is obtained by computing $\hat{e}_{\Phi}(z,y) = L_{2}(h_{\Phi}(z),h_{\Phi}(y))$.}
    \label{fig:lsl_ed}
\end{figure}

\subsubsection{LS-ED architecture.}
Char-CNN architecture~\cite{DBLP:conf/nips/ZhangZL15} is used for learning the deep embedding $h_{\Phi}$. It consists of five $1D$ convolution layers equipped with LeakyReLU activation \cite{DBLP:journals/corr/XuWCL15} followed by two fully connected layers. The embedding $h_{\Phi}$ maps the input such that $h_{\Phi}: \mathbb{R}^{|A|\times L} \xrightarrow{} \mathbb{R}^{1024}$. Note that since $h_{\Phi}$ constitutes of convolution and fully-connected layers, it is differentiable and allows for backpropagation to the STR model. In feed-forward, the two embeddings for the ground-truth $y$ (one-hot) and the model prediction $z$ (softmax) are obtained by performing feed-forward through $h_{\Phi}$ and an approximate of edit distance is computed by measuring the $L_{2}$ between the two vectors (Figure \ref{fig:lsl_ed}).

\subsubsection{Post-tuning with LS-ED.}
A random generator is designed for this task, which generates a pair of words $(z_{r}, y_{r})$ and ensures uniform sampling in the range of the true error. It was observed that the uniform sampling is essential to avert over-fitting of the learned surrogate on a certain range of the true metric. For the edit distance metric $e(z,y) \in \{0,...,b\}$ ($b$ being the maximum possible value), the generator samples a word randomly from a text corpus and distorts the words by performing random addition, deletion, and substitution operations.

The post-tuning of the STR model $f_{\Theta}(x)$ with LS-ED follows Algorithm~\ref{alg:train_with_lsl}. For the case of the edit distance, there is a significant domain gap between the samples obtained from the STR model ($z$) and the random generator ($z_{r}$). This is because the random generator operates directly on the text string, \textit{i.e.}, $z_{r}$ is one-hot representation. Thus, using the global approximation setting yields a low quality of the approximation. Further, it was observed that training the surrogate purely with the data generated from the STR model, \emph{i.e.}, local approximation, leads to a good approximation but does not lead to an improvement in the performance of the STR model, which indicates a low quality of gradients.

Finally as described in Algorithm \ref{alg:train_with_lsl}, the local-global approximation is used. The quality of approximation and the gradient penalty from post-tuning with LS-ED are shown in Figure \ref{fig:ls_ed_plots}. Note that the edit distance value is a whole number and the surrogate attempts to approximate it, thus the error in approximation as shown in Figure \ref{fig:ls_ed_plots} is low. The quality of the gradients can be seen by improvement in the performance of the STR models. Thus the local-global approximation guides to a high quality of both the approximation and gradients.

The results for the two configurations of STR models, \emph{i.e.}, \textit{ResNet-BiLSTM-Attn} and \textit{TPS-ResNet-BiLSTM-Attn}, are shown in Table \ref{table:resnet_bilstm_attn} and Table \ref{table:tps_resnet_bilstm_attn}, respectively. It can be observed that LS-ED improves the performance of the STR models on all metrics. The most significant gains are observed on total-edit distance (TED) as the surrogate attempts to minimize its approximation.

\begin{table}[t!]
\begin{center}
\begin{tabular}{ c | c | l | l | l}
\toprule
\textbf{\specialcell{Test \\ Data}} & \textbf{\specialcell{Loss \\ Function}} & \textbf{$\uparrow$ \specialcell{Acc.}} & 
\textbf{$\uparrow$ \specialcell{NED}} &
\textbf{$\downarrow$ \specialcell{TED}}\\
\midrule
IIIT-5K  & Cross-Entropy & $84.300$ & $0.954$ & $945$\\
IIIT-5K  & LS-ED  & $86.300$ \textcolor{green}{$+2.37\%$} & $0.953$ \textcolor{red}{$-0.10\%$} &  $837$ \textcolor{green}{$+11.42\%$}\\
\midrule
SVT  & Cross-Entropy & $84.699$ & $0.940$ & $229$\\
SVT  & LS-ED  & $86.399$ \textcolor{green}{$+2.00\%$} & $0.947$ \textcolor{green}{$+0.74\%$} & $196$ \textcolor{green}{$+14.41\%$}\\
\midrule
ICDAR'03 & Cross-Entropy & $92.558$ & $0.972$ & $151$\\
ICDAR'03 & LS-ED  & $94.070$ \textcolor{green}{$+1.63\%$} & $0.977$ \textcolor{green}{$+0.51\%$} & $119$ \textcolor{green}{$+26.89\%$}\\
\midrule
ICDAR'13 & Cross-Entropy & $89.754$ & $0.949$ & $260$ \\
ICDAR'13 & LS-ED  & $91.133$ \textcolor{green}{$+1.53\%$}& $0.960$ \textcolor{green}{$+1.15\%$}& $157$ \textcolor{green}{$+39.61\%$}\\
\midrule
ICDAR'15 & Cross-Entropy & $71.452$ & $0.889$ & $1135$ \\
ICDAR'15 & LS-ED  & $74.655$ \textcolor{green}{$+4.48\%$} & $0.899$ \textcolor{green}{$+1.12\%$} & $1013$ \textcolor{green}{$+10.74\%$}\\
\midrule
SVTP & Cross-Entropy & $74.109$ & $0.891$ & $424$ \\
SVTP & LS-ED  & $77.519$ \textcolor{green}{$+4.60\%$} & $0.901$ \textcolor{green}{$+1.22\%$} & $381$ \textcolor{green}{$+10.14\%$} \\
\midrule
CUTE & Cross-Entropy & $68.293$ & $0.838$ & $285$ \\
CUTE & LS-ED  & $71.777$ \textcolor{green}{$+5.10\%$} & $0.868$ \textcolor{green}{$+3.57\%$} & $234$ \textcolor{green}{$+17.89\%$}\\
\midrule
\end{tabular}
\end{center}
\caption{
ResNet-BiLSTM-Attn: The models are evaluated on IIIT-5K \cite{mishra2012scene}, SVT \cite{wang2011end}, ICDAR'03 \cite{lucas2003icdar}, ICDAR'13 \cite{karatzas2013icdar}, ICDAR'15 \cite{karatzas2015icdar}, SVTP \cite{quy2013recognizing}  and CUTE \cite{risnumawan2014robust} datasets. The results are reported using accuracy \textbf{Acc.} (higher is better), normalized edit distance \textbf{NED} (higher is better) and total edit distance \textbf{TED} (lower is better).
Relative gains are shown in \textcolor{green}{green} and relative declines in \textcolor{red}{red}.
}
\label{table:resnet_bilstm_attn}
\end{table}

\begin{table}[t!]
\begin{center}
\begin{tabular}{ c | c | l | l | l}
\toprule
\textbf{\specialcell{Test \\ Data}} & \textbf{\specialcell{Loss \\ Function}} & \textbf{$\uparrow$ \specialcell{Acc.}} & 
\textbf{$\uparrow$ \specialcell{NED}} &
\textbf{$\downarrow$ \specialcell{TED}}\\
\midrule
IIIT-5K  & Cross-Entropy & $87.500$ & $0.961$ & $722$\\
IIIT-5K  & LS-ED  & $87.933$ \textcolor{green}{$+0.49\%$} & $0.963$ \textcolor{green}{$+0.20\%$} & $645$  \textcolor{green}{$+10.66\%$}\\
\midrule
SVT  & Cross-Entropy & $87.172$ & $0.952$ & $180$\\
SVT  & LS-ED  & $86.708$ \textcolor{red}{$-0.53$} & $0.954$ \textcolor{green}{$+0.21\%$} & $163$ \textcolor{green}{$+9.44\%$}\\
\midrule
ICDAR'03 & Cross-Entropy & $94.302$ & $0.979$ & $110$\\
ICDAR'03 & LS-ED  & $94.535$ \textcolor{green}{$+0.24\%$} & $0.981$ \textcolor{green}{$+0.20\%$} & $99$       \textcolor{green}{$+10.00\%$}\\
\midrule
ICDAR'13 & Cross-Entropy & $92.020$ & $0.966$ & $137$ \\
ICDAR'13 & LS-ED  & $92.299$ \textcolor{green}{$+0.30\%$} & $0.979$ \textcolor{green}{$+1.34\%$} & $108$ \textcolor{green}{$+21.16\%$} \\
\midrule
ICDAR'15 & Cross-Entropy & $78.520$ & $0.915$ & $868$ \\
ICDAR'15 & LS-ED  & $78.410$ \textcolor{red}{$-0.14\%$} & $0.915$ \textcolor{black}{$\pm0.00\%$} & $837$ \textcolor{green}{$+3.57\%$} \\
\midrule
SVTP & Cross-Entropy & $78.605$ & $0.912$ & $346$ \\
SVTP & LS-ED  & $79.225$ \textcolor{green}{$+0.78\%$} & $0.913$ \textcolor{green}{$+0.10\%$} & $333$ \textcolor{green}{$+3.75\%$}\\
\midrule
CUTE & Cross-Entropy & $73.171$ & $0.871$ & $224$ \\
CUTE & LS-ED  & $74.216$ \textcolor{green}{$+1.42\%$} & $0.875$ \textcolor{green}{$+0.45\%$} & $219$ \textcolor{green}{$+2.23\%$}\\
\bottomrule
\end{tabular}
\end{center}
\caption{
TPS-ResNet-BiLSTM-Attn: The models are evaluated on IIIT-5K \cite{mishra2012scene}, SVT \cite{wang2011end}, ICDAR'03 \cite{lucas2003icdar}, ICDAR'13 \cite{karatzas2013icdar}, ICDAR'15 \cite{karatzas2015icdar}, SVTP \cite{quy2013recognizing}  and CUTE \cite{risnumawan2014robust} datasets.
The results are reported using accuracy \textbf{Acc.} (higher is better), normalized edit distance \textbf{NED} (higher is better) and total edit distance \textbf{TED} (lower is better).
Relative gains are shown in \textcolor{green}{green} and relative declines in \textcolor{red}{red}.
}
\label{table:tps_resnet_bilstm_attn}
\end{table}

\begin{table}
\begin{center}
\begin{tabular}{ l | l | l | l}
\toprule
\textbf{\specialcell{Loss \\ Function}} & \textbf{$\uparrow$ \specialcell{Recall}} & \textbf{$\uparrow$ \specialcell{Precision}} & \textbf{$\uparrow$ \specialcell{$F_{1}$ score}}\\
\midrule
 {\em Smooth-$L_{1}$}  & $71.21\%$ & $84.71\%$ & $77.37\%$ \\
 LS-IoU (global) & $66.97\%$ \textcolor{red}{$-5.95\%$} & $84.71\%$ \textcolor{black}{$\pm 0.00\%$} & $74.81\%$ \textcolor{red}{$-3.30\%$}\\
 LS-IoU (local) & $70.92\%$ \textcolor{red}{$-0.40\%$} & $86.60\%$ \textcolor{green}{$+2.23\%$} & $77.98\%$ \textcolor{green}{$+0.78\%$}\\
 LS-IoU (local-global) & $76.79\%$ \textcolor{green}{$+7.83\%$} & $84.93\%$ \textcolor{green}{$+0.25\%$} & $80.66\%$ \textcolor{green}{$+4.25\%$}\\
\bottomrule
\end{tabular}
\end{center}
\caption{RRPN-ResNet-50 \cite{ma2018arbitrary,ma2019rrpn}: Evaluations on Incidental Scene Text ICDAR'15 \cite{karatzas2015icdar}. Relative gains are shown in \textcolor{green}{green} and relative declines in \textcolor{red}{red}.}
\label{table:detection_results}
\end{table}

\subsection{Post-Tuning with a Learned Surrogate for IoU (LS-IoU)}
\label{sec:exp_iou}

It is experimentally demonstrated that LS can optimize scene text detection models on intersection-over-union (IoU) for rotated bounding boxes. IoU is a popular metric used to evaluate the object detection \cite{redmon2016you,ren2015faster} and scene text detection models \cite{ma2018arbitrary,buvsta2018e2e,DBLP:conf/cvpr/LiuLYCQY18,karatzas2015icdar,DBLP:conf/icdar/GomezSGNVMBK17}. Gradients for IoU can be hand-crafted for the case of axis-aligned bounding boxes \cite{yu2016unitbox,rezatofighi2019generalized}, however, it is complex to design the gradients for rotated bounding boxes. The learned surrogate of IoU allows backpropagation for rotated bounding boxes. For the task of rotated scene text detection on ICDAR'15 \cite{karatzas2015icdar}, it is shown that post-tuning the text detection model with LS-IoU leads to improvement on recall, precision, and $F_{1}$ score.

\subsubsection{Scene Text Detection.} 
Given a natural scene image, the objective is to obtain precise word-level rotated bounding boxes. The method proposed by Ma~\etal~\cite{ma2018arbitrary} is used for the task. It extends Faster-RCNN \cite{ren2015faster} based object detector to incorporate rotations. This is achieved by adding angle priors in anchor boxes to enable rotated region proposals. A sampling strategy using IoU compares these proposals with the ground truth and filter the positive and the negative proposals. Only the filtered proposals are used for the loss computation.

The positive proposals are regressed to fit precisely with the ground truth. Through rotated region-of-interest (RROI) pooling, the features corresponding to the proposals are obtained and used for text/no-text binary classification. The overall loss function for training in \cite{ma2018arbitrary} is defined as a linear combination of classification loss (negative log-likelihood) and regression loss ({\em smooth-L$_{1}$}).

The publicly available implementation of \cite{ma2018arbitrary,ma2019rrpn} is used with the original hyper-parameter settings -- the model is trained for $140K$ iterations using the SGD optimizer and batch-size of $1$. The model is trained on a union of ICDAR'15 \cite{karatzas2015icdar} and ICDAR-MLT \cite{nayef2019icdar2019} datasets, providing $6295$ training images.

\subsubsection{LS-IoU architecture.} 
The embedding model for LS-IoU consists of five fully-connected layers with ReLU activation \cite{DBLP:journals/jmlr/GlorotBB11}. A rotated bounding box is represented with six parameters, two for the coordinates of the centre of the box, two for the height and the width and two for {\em cosine} and {\em sine} of the rotation angle. The centre coordinates and the dimensions of the box are normalized with image dimensions to make the representation invariant to the image resolution.

The embedding model maps the representation of a positive box proposal and the matching ground-truth into a vector as $h_{\Phi}: \mathbb{R}^{6}\xrightarrow{} \mathbb{R}^{16}$. The approximation of the IoU between two bounding boxes is computed by the $L_{2}$ distance between the two vector representations.

\subsubsection{Post-tuning with LS-IoU.} 
The random generator for LS-IoU samples rotated bounding boxes from the set of training labels and modifies the boxes by changing the centre locations, dimensions, and rotation angle within certain bounds to create a distorted variant. Since uniform sampling over the range of IoU is difficult, we store roughly $3$ million such examples along with the IoU values and sample from this collection.

Note that since the overall loss for training \cite{ma2018arbitrary} is a combination of a regression loss and a classification loss, LS-IoU only replaces the regression component ({\em smooth-$L_{1}$}) with the learned surrogate for IoU. For post-tuning with LS-IoU, the results are shown for all three setups, that is, global approximation, local approximation and global-local approximation (Algorithm \ref{alg:train_with_lsl}). For each of these, the model trained with proxy losses is post-tuned with LS-IoU for $20K$ iterations. The quality of the approximations for the first $1K$ iterations of the training is shown in Figure \ref{fig:ls_iou_plots}. Since the range of IoU is in $[0,1]$, it can be seen that the error is high for the global approximation. For both local and global-local, the quality of the approximation is significantly better (roughly $10$ times lower error).

As mentioned earlier, the quality of gradients is judged by the improvement or deterioration of the model ($f_{\Theta}(x)$) post-tuned with LS-IoU. The results for scene text detection on the ICDAR'15 \cite{karatzas2015icdar} dataset are shown in Table \ref{table:detection_results}. It is observed that post-tuning the detection model with LS-IoU (global) leads to deterioration. Post-tuning with LS-IoU (local) improves the precision but makes recall worse. Finally, LS-IoU (local-global) from Algorithm \ref{alg:train_with_lsl} improves both the precision and recall, boosting the $F_{1}$ score by relative $4.25\%$.

\section{Conclusions}
\label{sec:conclusion}

A technique is proposed for training neural networks by minimizing learned surrogates that approximate the target evaluation metric. The effectiveness of the proposed technique has been demonstrated in a post-tuning setup, where a trained model is tuned on the learned surrogate. Improvements have been achieved on the challenging tasks of scene-text recognition and detection. By post-tuning, the model with LS-ED, relative improvements of up to $39\%$ on the total edit distance has been achieved. On detection, post-tuning with LS-IoU has shown to provide a relative gain of $4.25\%$ on the $F_{1}$ score.

\section*{Acknowledgement}
The authors thank R. Manmatha, Dmytro Mishkin, Michal Bu{\v{s}}ta, Kl{\'a}ra Janou{\v{s}}\-ko\-v{\'a}, Viresh Ranjan and Abhijeet Kumar for the feedback. This research was supported by Research Center for Informatics (project CZ.02.1.01/0.0/0.0/\-16019/0000765 funded by OP VVV) and CTU student grant (SGS OHK3-019/20).

\bibliographystyle{splncs04}
\bibliography{egbib}
\end{document}